\documentclass{article}
\usepackage{float}

\usepackage[preprint]{neurips_2026}

\usepackage[utf8]{inputenc} 
\usepackage[T1]{fontenc}    
\usepackage{hyperref}       
\usepackage{xurl}            
\usepackage{enumitem}
\usepackage{booktabs}       
\usepackage{amsfonts}       
\usepackage{amsmath}
\usepackage{nicefrac}       
\usepackage{microtype}      
\usepackage{xcolor}         
\usepackage{graphicx}
\usepackage{multirow}
\usepackage[section]{placeins}

\hypersetup{pdfborder={0 0 0}} 
\setlist[itemize]{leftmargin=*}
\setlist[enumerate]{leftmargin=*}

\title{Kuration SDK: Addressing the Virtual2Real Gap via Data Curation}
\workshoptitle{Robot Learning with World Models: Capabilities, Frontiers, and Challenges}
\author{
  \textbf{Nirmit Desai\thanks{Corresponding author: \texttt{nirmit.desai@aintropy.ai}}, Eric Song, Mayank Sengupta, Tejal Bedmutha,} \\
  \textbf{Siri Reddy, Sahiti Dharmavaram, Kunal Sawarkar} \\
  AIntropy AI
}

\begin{document}

\maketitle

\begin{abstract}
Benchmarks for measuring the quality of action-conditioned world models are still evolving and shifting away from visual similarity-based metrics to action-semantic and physically-grounded metrics. However, for domain and task-agnostic action-conditioned world model training, existing benchmarks provide a limited signal. By training and evaluating diffusion world models on CounterStrike gameplay data, we confirm that qualitative playability does not correspond with metrics such as FVD, LPIPS, and JEDi. We term this the \emph{Virtual2Real} gap.
We posit that, in lieu of reliable benchmarks, curating raw gameplay data and measuring a variety of diagnostic properties provides a more robust signal to bridge the gap, before the training even begins. We present several curation strategies and a general-purpose kit for physical AI data curation called \emph{Kuration SDK}, which is being open-sourced with this paper \footnote{https://drive.google.com/file/d/1tbq9zW12YJTt9bw4UL2n3r576je1G0OU}. The SDK was instrumental in uncovering the root cause of the virtual2real gap in a specific case: why two world models trained on identical gameplay map, action and state distribution, behaved very differently when played in spite of having very similar LPIPS and FVD scores. Thus, Kuration SDK has the potential to uncover the root causes of Virtual2Real gap in specific datasets and accelerate development of sample-efficient training datasets. 
\end{abstract}

\section{Introduction}
\label{sec:intro}
In the case of LLMs, training data curation and data quality ablation experiments were critical in accelerating task performance of LLMs, e.g., large-scale web-text curation pipelines such as FineWeb~\citep{fineweb} that filter and deduplicate raw crawl data before it ever reaches a training run. Well-established and meaningful benchmark metrics for evaluating foundational language capability enabled that progress, and generative modeling more broadly has shown predictable improvements with more data and compute~\citep{henighan_scaling}. However, meaningful metrics for action-conditioned world models are still evolving. State-of-the-art benchmarks are either task-specific and require an agent to be trained and evaluated on task performance or do not provide a meaningful signal on model capability. Then, what is the role of data curation for training foundational and task-agnostic action-conditioned world models? This is the central question this paper investigates.

Playing a pretrained DIAMOND~\citep{diamond} model, a diffusion world model trained on CS:GO gameplay, makes the problem concrete. Rollouts are visually coherent for roughly the first thirty frames; beyond that horizon, the model loses track of the player's position and teleports to a visually similar but geographically distinct part of the map, and it permits actions the game physics never allow, such as walking through parked cars or repeated double-jumps against a wall. We use \emph{playability} throughout to mean exactly these observable failures, map teleportation, geometry clipping, physics-violating actions, and control responsiveness and recovery, judged by the authors via interactive rollout inspection rather than a formal protocol or external, blinded raters (Section~\ref{sec:limitations}). Such failures are attributed to the training data rather than the architecture: raw competitive gameplay rarely contains sustained collisions, since good players avoid walking into walls, and it contains double-jump exploits often enough for the model to treat them as normal transitions.

By training action-conditioned world models on two gameplay datasets having identical state and action distributions, identical game world map, identical volumes, frame rates, resolution, and near identical game engines (CounterStrike:GO and CounterStrike2), we show that state-of-the-art benchmarks do not correspond to qualitative "playability" of the models. Similar to the Sim2Real gap, we term this the \emph{Virtual2Real} gap. Given the lack of meaningful benchmarks to investigate this gap, we built a set of data diagnostics and curation tools we call \emph{Kuration SDK}. The SDK was instrumental in uncovering the root cause of the gap: the magnitude of visual shift per action differed by $1.4$--$2.1\times$ across the two datasets. Section~\ref{sec:related} situates this against prior world-model and benchmark work; Sections~\ref{sec:baseline}--\ref{sec:asc} present the controlled experiment and root-cause diagnosis; Section~\ref{sec:results} evaluates curation strategies against FVD, LPIPS, and JEDi; Section~\ref{sec:conclusion} discusses the implications for benchmarking gameplay world models. Our contributions:
\subsection{Contributions}
\begin{itemize}
\item We demonstrate the Virtual2Real gap directly: two diffusion world models trained on format- and distribution-matched CS:GO and CS2 corpora diverge sharply in qualitative playability despite comparable or better scores on FVD, LPIPS, and JEDi.
\item We survey emerging action-semantic and physically-grounded benchmarks and show none offers a general-purpose, domain-agnostic quality signal for gameplay world models, motivating data-level diagnostics as an alternative.
\item Using the Kuration SDK, we test a set of curation hypotheses and identify action-state consistency, whether a given input reliably produces the same characteristic screen motion, and specifically its magnitude, as the leading candidate explanation: the two corpora differ by roughly $2\times$ on this axis, a difference neither FVD nor LPIPS detects.
\item We release the Kuration SDK, an open-source toolkit for gameplay data curation, together with the curation strategies and full evaluation results from this paper, so the community can reproduce, extend, or apply these diagnostics to new corpora.
\end{itemize}

\section{Related work}
\label{sec:related}

\paragraph{World models for physical AI.} World models predicting future observations conditioned on actions have been proposed as planning and policy-learning substrates without further real-world interaction~\citep{world_models_ha,dreamerv3}. Genie~\citep{genie} pursues data scale, pretraining on internet-scale video with no action labels. GameNGen~\citep{gamengen} and DIAMOND~\citep{diamond}, real-time diffusion game world models, are the closest comparisons; we use DIAMOND as our baseline. Reported failure modes (map teleportation, geometry clipping) are attributed to the training corpus, but, to our knowledge, no prior work tests this against a second, independently captured corpus matched on format and scale, the gap Section~\ref{sec:cs2} closes.

\paragraph{Sim2Real vs.\ Virtual2Real.} Sim2Real~\citep{sim2real,tobin2017} treats the gap as dynamics-fidelity, addressed via domain randomization. Virtual2Real is a data-distribution problem: a model trained on an uncurated virtual-environment sample inherits its degenerate corners regardless of simulator accuracy.

\paragraph{Video generation metrics and their limits.} FVD~\citep{fvd} and LPIPS~\citep{lpips} compute a Fr\'echet distance over I3D features and a per-frame perceptual distance; neither targets single-interaction physical violations. JEDi~\citep{jedi} replaces I3D with V-JEPA~\citep{vjepa1} features, predicting future latent states rather than classifying actions, and so is, in principle, more sensitive to how a scene evolves (Sections~\ref{sec:cs2},~\ref{sec:results}). Newer benchmarks targeting action-semantics or physical grounding directly are not yet general-purpose: ACT-Bench~\citep{actbench} needs a per-game trajectory or pose estimator, built for continuous 3D ego-motion rather than discrete or hybrid action spaces; ARR~\citep{arr_multiplayer_wm} depends on an auxiliary probe trained and validated per game; Atari~100K~\citep{atari100k} and similar task-specific benchmarks do not generalize; and long-horizon drift metrics~\citep{stable_world_models}, an action sequence followed by its inverse checked for return to the start, are self-referential, a round trip can close on an arbitrarily wrong path since no step is individually verified.

\paragraph{Data curation.} Subset selection from a large corpus is well studied for language and vision-language data (semantic deduplication~\citep{semdedup}, coverage-driven filtering~\citep{dclm}) over fixed text or image embeddings, which do not expose the features that matter here, physical interaction, action-state consistency, spatial trajectory, thus motivating gameplay-specific operators.

\section{Establishing a training baseline: reproducing DIAMOND}
\label{sec:baseline}

To attribute any downstream effect to data rather than to our own training or evaluation code, we first reproduced DIAMOND on its published CS:GO dataset.

\paragraph{Architecture and setup.} DIAMOND is a two-stage architecture: a denoiser conditioned on the previous four frames and actions, operating at $30\times56$, and an independently trained upsampler mapping denoiser output to $150\times280$. Actions are discrete, spanning 20 categories of mouse and keyboard input. We matched the published configuration (learning rate $10^{-4}$, weight decay $10^{-2}$, 600 epochs of 400 steps), and effective batch size.

\paragraph{Parity result.} Our 600-epoch run reaches FVD $4.4\%$ below DIAMOND's reported value at epoch 300 and LPIPS $4.2\%$ above it at epoch 420, both within evaluation stochasticity. Running the released DIAMOND weights through our own pipeline reproduces the paper's numbers within $1\%$, confirming our evaluation code (Table~\ref{tab:parity}; per-30-epoch curve in Figure~\ref{fig:lpips_fvd}, Appendix~\ref{app:full_results}). Interactive rollouts from the reproduced model match DIAMOND's published qualitative standard at short horizons; the long-horizon spatial teleportation and geometry-clipping failure modes described in Section~\ref{sec:intro} are present but, being multi-frame phenomena, are not visible in any single frame, part of why frame-level and short-clip metrics struggle to detect them.

\begin{table}[t]
\centering
\caption{Training parity with DIAMOND on the published CS:GO dataset. All evaluations use 500 test episodes and 1-step samplers.}
\label{tab:parity}
\begin{tabular}{llcc}
\toprule
Run & Epoch & LPIPS $\downarrow$ & FVD $\downarrow$ \\
\midrule
DIAMOND (paper, reported) & 600 & $0.5408 \pm 0.1829$ & $1{,}643{,}391$ \\
DIAMOND weights (our eval) & 600 & $0.5414 \pm 0.1836$ & $1{,}655{,}984$ \\
\midrule
Ours (best FVD) & 300 & $0.5789 \pm 0.1839$ & $\mathbf{1{,}571{,}046}$ \\
Ours (best LPIPS) & 420 & $\mathbf{0.5634 \pm 0.1916}$ & $1{,}801{,}407$ \\
\bottomrule
\end{tabular}
\end{table}

The reproduction confirms two things before we touch curation: our training and evaluation pipeline is faithful, and it reproduces the published qualitative failures (Section~\ref{sec:intro}), map teleportation and geometry clipping persist at the same visual standard as the original. Any difference observed once we swap in a new corpus (Section~\ref{sec:cs2}) can therefore be attributed to the data.

\section{Controlled replication: Training on partner CS2 data}
\label{sec:cs2}
We obtained server-side-captured Counter-Strike~2 (CS2) gameplay from a data partner, matched to DIAMOND's CS:GO corpus on map (\texttt{de\_dust2}), capture rate (16~fps), action encoding, and corpus scale; the recorded content, which players, which matches, which moments, is the one variable that cannot be held fixed across an independently captured corpus.

We trained an uncurated reference model, \texttt{baseline\_v1}, on $8{,}635$ raw CS2 episodes, identical architecture and hyperparameters to Section~\ref{sec:baseline}, no episode-level curation, evaluated on a held-out set of $1{,}090$ CS2 episodes. Cross-evaluating DIAMOND's weights on this CS2 test set (Table~\ref{tab:strategies}) gives LPIPS and FVD both markedly worse than its own native CS:GO-test-set scores (Table~\ref{tab:parity}), a real domain gap despite the shared map and format.

\texttt{baseline\_v1} improves LPIPS by $20.9\%$ over DIAMOND on the CS2 test set by epoch 600, matching or exceeding DIAMOND on aggregate video-generation metrics (Table~\ref{tab:strategies}). It is nonetheless qualitatively \emph{less} playable: in interactive rollouts it shows the same class of failures as CS:GO DIAMOND, geometry clipping and spatial discontinuity over longer horizons, more severely, with less responsive control and less clean recovery. FVD is also a poor stopping signal here: its best value occurs at epoch 60 and is not bettered over the remaining $540$ epochs (Appendix~\ref{app:full_results}). Cross-evaluated back onto DIAMOND's own test set, \texttt{baseline\_v1} scores FVD $4.7\times$ worse than DIAMOND: it does not transfer even across two versions of the same game on the same map.

Table~\ref{tab:strategies} (Section~\ref{sec:results}) reports \texttt{baseline\_v1}'s full metrics alongside DIAMOND and every curation strategy on the same test set.

\begin{figure}[H]
  \centering
  \includegraphics[width=0.75\linewidth]{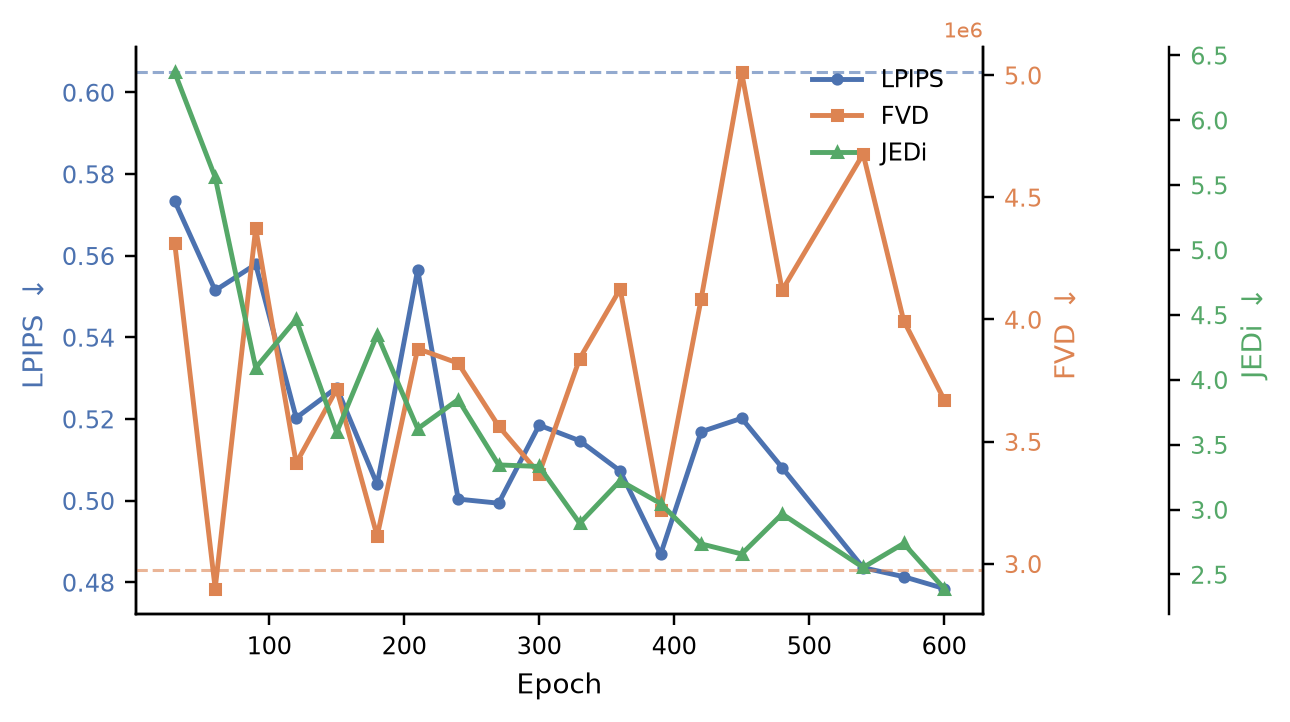}
  \caption{LPIPS, FVD, and JEDi for \texttt{baseline\_v1} across training, evaluated on the CS2 test set, plotted on three shared axes. Dashed lines show the DIAMOND model on the same test set (no JEDi reference exists on this test set). LPIPS improves with noise; FVD is competitive only at early epochs and then oscillates; JEDi improves at every evaluated checkpoint but playability is still worse.}
  \label{fig:cs2_metrics}
\end{figure}

\section{Action-State consistency: the signal behind the gap}
\label{sec:asc}

Section~\ref{sec:cs2}'s two corpora agree on map, resolution, frame rate, and action encoding, yet produced models with different specialization and the same class of playability failure. We formulate several hypotheses and evaluate them via Kuration SDK as shown later in Section~\ref{sec:results}. Of those hypotheses, the \texttt{action\_state\_consistency} hypothesis was validated as described here.

\paragraph{The operator.} An action-conditioned world model learns only the mapping between an input and the screen motion that follows it; if that mapping is inconsistent in the data, the model cannot distinguish a genuine rule from an anomaly. Game engines routinely break this contract for reasons unrelated to world dynamics, coyote time, input buffering, lag compensation, deliberate choices that make a game feel responsive to a human player. \texttt{pixel\_shift} measures the optical-flow displacement following every uninterrupted key press, projected onto the axis that control should move the camera on; \texttt{action\_state\_consistency} aggregates this into a per-control, per-press-length response curve.

\begin{figure}[t]
  \centering
  \includegraphics[width=0.85\linewidth]{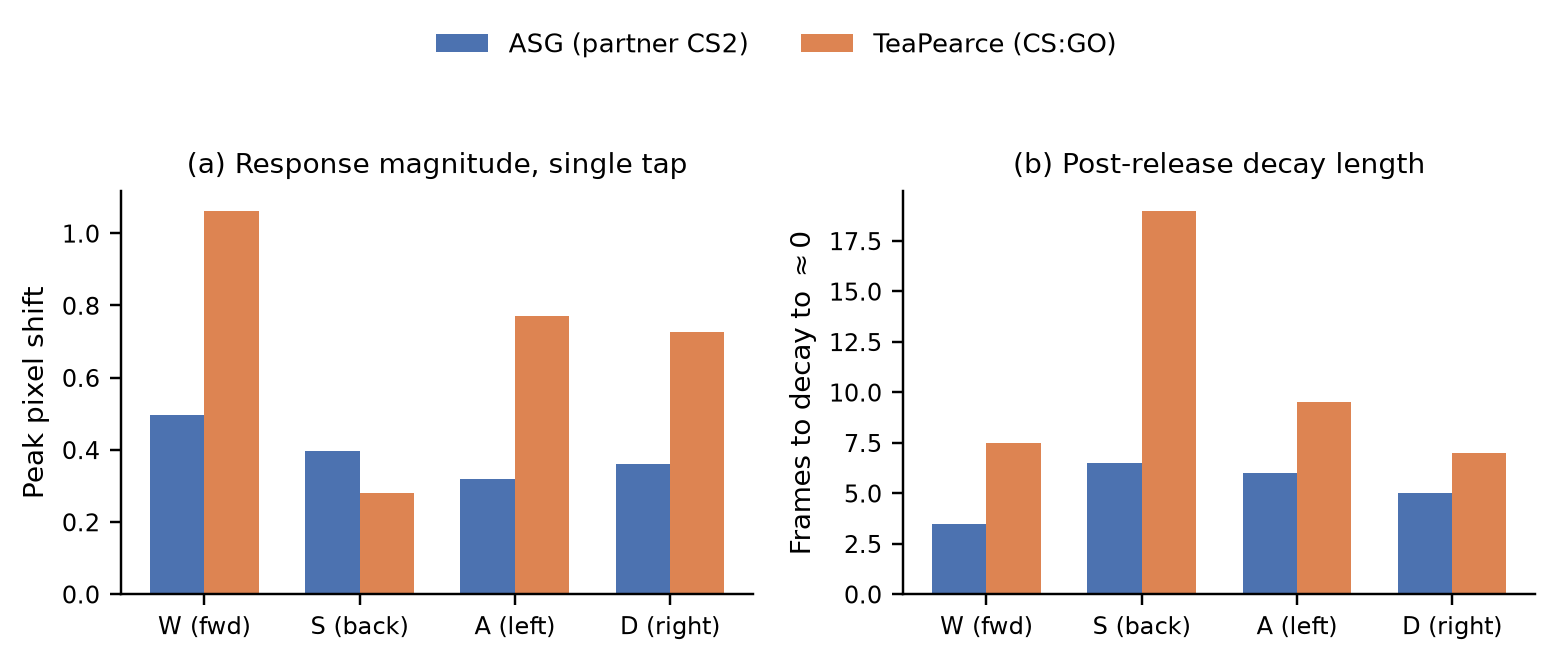}
  \caption{Action-state consistency for single key-tap movement controls, DIAMOND's public CS:GO corpus vs.\ the partner CS2 corpus. (a) CS:GO responses are consistently larger in magnitude. (b) CS:GO responses take substantially longer to decay to zero after release, evidence of sustained momentum absent from the CS2 corpus.}
  \label{fig:asc}
\end{figure}

\paragraph{Finding.} Applying this to DIAMOND's CS:GO corpus~\citep{diamond} and the partner CS2 corpus shows the two are not interchangeable at the action level, despite matching every specification and distribution. Figure~\ref{fig:asc} summarizes single-tap responses for the four movement controls. Three differences stand out. \textbf{Magnitude}: CS:GO responses are $1.4$--$2.1\times$ larger in peak pixel shift across all four controls. \textbf{Momentum}: on multi-frame holds, CS2 decelerates on release ($76$--$78\%$ of in-hold magnitude, matching physical expectation), while CS:GO accelerates through release ($112$--$126\%$) and takes $2$--$3\times$ longer to decay to zero. \textbf{Symmetry}: CS2's backward (S) response is $80\%$ of forward (W), close to physically expected near-symmetry; CS:GO's is only $26\%$.

\paragraph{Why this matters.} This is a measurable difference neither FVD nor LPIPS can detect. It also tracks which model plays better: CS:GO DIAMOND, despite its own map-teleportation and clipping failures, is more responsive and controllable than \texttt{baseline\_v1}, despite \texttt{baseline\_v1} matching or beating it on LPIPS and FVD. Of the three axes, magnitude is the most plausible driver: a larger, more visually salient pixel shift per action gives a denoising, action-conditioned model a stronger training signal to key its predictions on, relative to other sources of frame-to-frame variation; a weaker signal, as in CS2, is easier to under-weight, even though CS2 is arguably the more physically well-behaved corpus on momentum and symmetry. We have not yet run a controlled ablation isolating magnitude from momentum and symmetry, so we report magnitude as the leading hypothesis, not a proven one. CS:GO and CS2 also run on different engine generations (Source vs.\ Source~2), so part of this measured difference could in principle reflect an engine-level property rather than a purely curatable data-selection one; matching map, resolution, frame rate, and action encoding (Section~\ref{sec:cs2}) narrows this confound but does not eliminate it (Section~\ref{sec:limitations}). What is established without qualification: two corpora indistinguishable by format specification differ measurably in exactly the property an action-conditioned model depends on, and that difference tracks playability where FVD and LPIPS do not.

\section{Kuration SDK}
\label{sec:sdk}

The action-state consistency finding above is one instance of a general capability: the \textbf{Kuration SDK}, an open-source, pip-installable toolkit we release with this paper for curating gameplay and other sequential data into training-ready shards.\footnote{See the abstract for the access link.} A pipeline windows a source into segments, measures per-segment features, filters or selects against a time or count budget under a leakage-free train/test split, and writes shards with a config, manifest, and checksum, so a curated dataset is exactly reproducible from its config alone.

\texttt{kuration.gameplay} ships the operators used in this paper. \texttt{pixel\_shift} and \texttt{action\_state\_consistency} implement Section~\ref{sec:asc} (pipeline example in Appendix~\ref{app:strategy_spec}). \texttt{collision} finds and classifies wall-hits from an expected unobstructed-movement curve so they can be oversampled relative to their naturally low base rate, targeting the wall-clipping failure mode of Section~\ref{sec:intro}. \texttt{slam\_reconstruct}, \texttt{slam\_quality}, and \texttt{slam\_motion\_diversity} recover a camera's 3D trajectory from video with no positional metadata and score it for gaps, impossible motion, and turning and path-length diversity, turning spatial coverage into a curatable axis for footage that started with none. Three general-purpose selection recipes, equal coverage across quantile bins of any feature, oversampling a rare event to a target rate, and cluster-then-sample diversity selection, compose with any of the above without new code. All strategies used below, and the full evaluation results, are released alongside the SDK.

\section{Curation strategies and results}
\label{sec:results}

The action-state consistency finding motivates curating for specific behavioral properties rather than simply collecting more data, but it was one of four hypotheses we tested, not a foregone conclusion. Using the SDK, we curated the CS2 corpus along four strategies and compared each against \texttt{baseline\_v1} as the uncurated control, judging every checkpoint by the same qualitative rollout inspection used in Section~\ref{sec:cs2}. \textbf{Coverage-based sampling} selects segments to uniformly cover quantile bins over aim velocity, input entropy, and movement speed, trained from scratch for 120 of a planned 600 epochs on 7,530 episodes. \textbf{Rare-trajectory up-weighting} selects segments where movement speed falls in the tail (mean $43.9$ vs.\ corpus mean $27.2$), fine-tuned 60 epochs from \texttt{baseline\_v1}'s epoch-300 checkpoint. \textbf{Collision-aware selection} up-weights collision-flagged segments (impact rate weighted $2\times$ movement speed, $4\times$ aim velocity), fine-tuned 55 epochs from the same checkpoint. \textbf{Action-state-consistency selection}, the only strategy motivated directly by the Section~\ref{sec:asc} diagnosis, up-weights segments with the largest per-action pixel-shift magnitude, moving the training distribution toward CS:GO's more salient action-to-motion contract.

\begin{table}[t]
\centering
\caption{DIAMOND and curation strategies against the uncurated \texttt{baseline\_v1} control, all evaluated on the CS2 test set (1,090 episodes). Coverage-based training is from scratch; rare-trajectory and collision-aware are fine-tuned from \texttt{baseline\_v1}'s epoch-300 checkpoint (see Appendix~\ref{app:full_results} for full per-epoch curves; Section~\ref{sec:no_correlation} for coverage-based's epoch-105 best checkpoint). Action-state-consistency's training details are not yet finalized (Section~\ref{sec:limitations}). Playability is a qualitative judgment from interactive rollouts, relative to \texttt{baseline\_v1}.}
\label{tab:strategies}
\resizebox{\linewidth}{!}{%
\begin{tabular}{lcccccc}
\toprule
Strategy & Epochs & Init.\ & LPIPS $\downarrow$ & FVD $\downarrow$ & JEDi $\downarrow$ & Playability \\
\midrule
DIAMOND (our eval) & 600 & scratch & $0.6049$ & $2{,}974{,}382$ & --- & better \\
\texttt{baseline\_v1} (uncurated) & 600 & scratch & $\mathbf{0.4785}$ & $\mathbf{2{,}898{,}653}^{\dagger}$ & $\mathbf{2.39}$ & reference \\
Coverage-based & 120 & scratch & $0.5404$ & $3{,}043{,}786$ & --- & no change \\
Rare-trajectory & 60 & epoch 300 & $0.5043$ & --- & $3.13$ & no change \\
Collision-aware & 55 & epoch 300 & --- & --- & $2.76$ & no change \\
Action-state-consistency & --- & --- & --- & --- & --- & \textbf{improved} \\
\bottomrule
\end{tabular}%
}
\vspace{2pt}
\footnotesize{$^{\dagger}$Best FVD occurs at epoch 60, not at \texttt{baseline\_v1}'s best-LPIPS checkpoint; the row mixes each strategy's best value per metric, not a single checkpoint, since that is how each metric would be used as a stopping criterion in practice. Bold marks the lowest (best) value per metric among the CS2 runs. Blank cells indicate the corresponding evaluation was not completed at the time of writing (see Section~\ref{sec:limitations}).}
\end{table}

\subsection{One strategy improves playability}
\label{sec:no_correlation}

Of the four curation strategies tested, three, coverage-based, rare-trajectory, and collision-aware, produce no qualitative playability improvement over \texttt{baseline\_v1}, despite each moving at least one metric in the direction that would ordinarily be read as progress: coverage-based reaches near-baseline LPIPS and FVD in a fifth of the training budget ($105$ of $120$ epochs), and collision-aware improves JEDi over the comparably fine-tuned rare-trajectory run (Table~\ref{tab:strategies}), consistent with collision-rich data helping a metric sensitive to interaction dynamics. Only action-state-consistency selection, the strategy directly motivated by the Section~\ref{sec:asc} root-cause diagnosis rather than a plausible-sounding proxy, produces a noticeable improvement in playability over \texttt{baseline\_v1}; we do not yet know whether it closes the gap to CS:GO DIAMOND or narrows it only partially, and we have not run the negative control of curating for the same axis in the wrong direction, so we cannot yet rule out that any sufficiently different fine-tune from the epoch-300 checkpoint would have moved playability. Curating along an axis that sounds physically relevant (rare, physically demanding trajectories; collision events) was not sufficient here, and metric movement on that axis was not a reliable sign that it worked either way.

This is not a claim that FVD and LPIPS are useless: LPIPS tracks progress within a single run, and FVD crosses below DIAMOND's reported value by epoch 180 of the reproduction (Figure~\ref{fig:lpips_fvd}), a coarse but real convergence signal. The claim is narrower: once models are individually competent, neither metric, nor JEDi, reliably flags which curation hypothesis actually fixes the targeted behavior, which is precisely the regime curation research operates in.

\subsection{Limitations}
\label{sec:limitations}

Several caveats. Collision-aware LPIPS/FVD, coverage-based JEDi, and all three of action-state-consistency's quantitative metrics were not completed at time of writing (shared GPU availability); blank cells in Table~\ref{tab:strategies} should not be read as parity, and the action-state-consistency result, including its training schedule, is qualitative only so far. The playability judgments throughout this paper rest on rollout inspection rather than a labeled playability benchmark, since none exists for this setting to our knowledge; a labeled benchmark, or even a second independent rater, would let us report agreement rather than a single team's judgment. Raters were the authors themselves and were not blinded to which strategy produced a given checkpoint, so expectation bias cannot be ruled out, a concern that applies most directly to the action-state-consistency result, the one motivated by our own root-cause diagnosis. Neither a systematic, shared-test-set sweep of the SDK's full strategy space nor a statistical action-state-consistency comparison with confidence intervals per control and press length has been run; Section~\ref{sec:asc} reports single-tap summaries, and multi-tap presses show the same pattern but are omitted here for space. CS:GO and CS2 are also not the same engine: beyond the map, resolution, frame rate, and action encoding we matched, they differ in engine generation (Source vs.\ Source~2) and underlying physics and netcode, so part of the action-state consistency gap of Section~\ref{sec:asc} could reflect an engine-level property rather than a purely fixable data-selection one; we cannot yet separate the two with a single corpus pair. More broadly, every result in this paper comes from one map, one architecture (DIAMOND), and one game family; the "task-agnostic" motivation in Section~\ref{sec:intro} is the question this work investigates, not a generality we have yet demonstrated beyond this setting.

\section{Conclusion and Future Work}
\label{sec:conclusion}

With the example of two format and scale-matched CounterStrike gameplay corpora, we demonstrated the Virtual2Real gap: benchmarks improve slightly while playability gets significantly worse. Faced with no reliable benchmark to identify the root cause, we used the Kuration SDK to test curation hypotheses directly against qualitative playability rather than against metrics alone (Section~\ref{sec:results}); only the hypothesis grounded in a measured difference between corpora, not an intuition about which behaviors seemed important, moved playability. In our experience, the SDK's value was in letting us test and discard hypotheses without one-off scripts for each, not in any single curation trick, though we have not measured this efficiency claim against an alternative workflow. We release the Kuration SDK, the curation strategy specifications, and the full evaluation results from this paper so other groups can reproduce, extend, or apply these diagnostics to new corpora.

In next steps, we will build a \textbf{physics-conformance metric} directly from geometry pass-through and trajectory discontinuity, rather than learned video features. Also, a \textbf{shared benchmark} comparing the SDK's full strategy space on one held-out set, against all metrics used here plus any physics-conformance metric will make the claims more conclusive. \textbf{Generalization beyond CS2}: whether the same curation axes (interaction density, spatial coverage, action-state consistency) transfer to real robot teleoperation or egocentric video would make the case for domain-general curation infrastructure concrete rather than analogical.

\bibliographystyle{plainnat}
\bibliography{references}

\newpage
\appendix

\section{Full Per-Checkpoint Results}
\label{app:full_results}

\begin{figure}[h]
  \centering
  \includegraphics[width=0.8\linewidth]{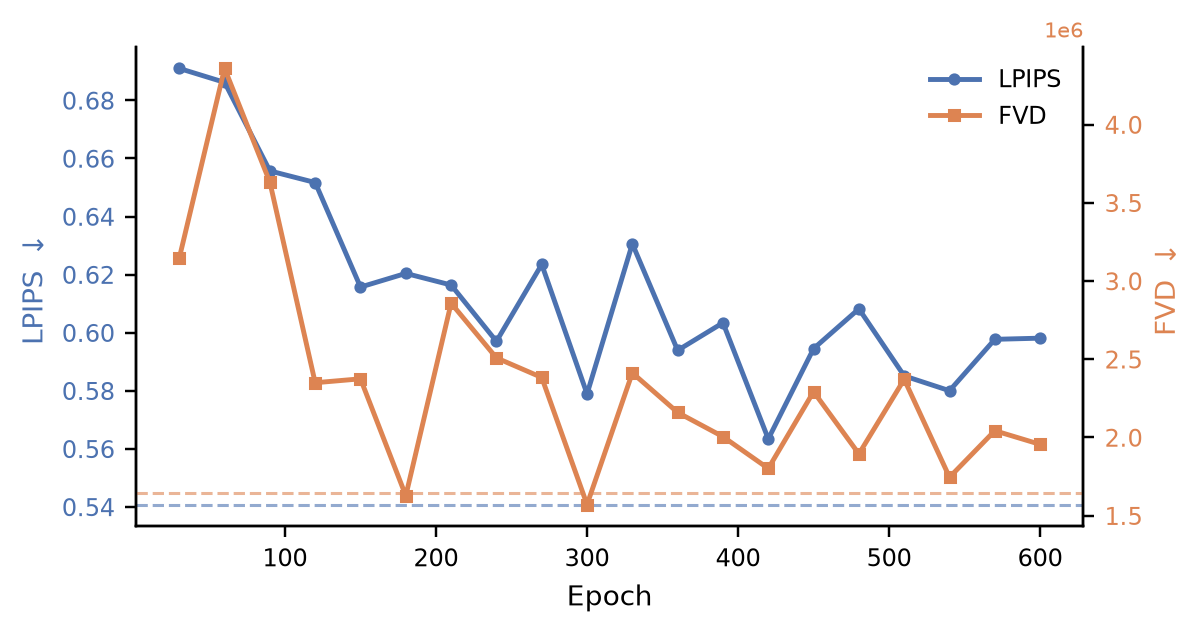}
  \caption{LPIPS (left axis) and FVD (right axis) at each 30-epoch checkpoint of our 600-epoch DIAMOND reproduction on CS:GO, plotted on a shared axis. Dashed lines mark DIAMOND's reported values. FVD crosses below the paper's reported value at epoch 180 and reaches its best value at epoch 300; after that, both metrics plateau and oscillate without a consistent trend even though training loss continues to decrease monotonically.}
  \label{fig:lpips_fvd}
\end{figure}

Table~\ref{tab:full_diamond} reports LPIPS, FVD, and JEDi at every 30-epoch checkpoint of our DIAMOND reproduction (Section~\ref{sec:baseline}), on the DIAMOND standard CS:GO test set (500 episodes, 1-step samplers), plotted in Figure~\ref{fig:lpips_fvd}. Denoiser train loss decreases monotonically from $0.0682$ at epoch 30 to $0.0339$ at epoch 600, and denoiser test loss (evaluated at every second checkpoint) decreases from $0.0469$ at epoch 60 to $0.0240$ at epoch 600, confirming that the model continues to improve on its core frame-prediction objective throughout training even as FVD and LPIPS plateau after epoch 300. JEDi on this run is noisier than on the CS2 run of Table~\ref{tab:full_cs2} and does not improve monotonically, though its best value is also reached late in training.

\begin{table}[h]
\centering
\caption{LPIPS, FVD, and JEDi for the DIAMOND reproduction at each 30-epoch checkpoint, DIAMOND standard CS:GO test set.}
\label{tab:full_diamond}
\begin{tabular}{cccc}
\toprule
Epoch & LPIPS $\downarrow$ & FVD $\downarrow$ & JEDi $\downarrow$ \\
\midrule
30  & $0.6910 \pm 0.1518$ & $3{,}145{,}978$ & $5.5917$ \\
60  & $0.6862 \pm 0.1634$ & $4{,}358{,}746$ & $4.9600$ \\
90  & $0.6557 \pm 0.1716$ & $3{,}631{,}473$ & $3.5018$ \\
120 & $0.6517 \pm 0.1749$ & $2{,}349{,}638$ & $2.9815$ \\
150 & $0.6157 \pm 0.1778$ & $2{,}374{,}627$ & $2.9877$ \\
180 & $0.6204 \pm 0.1823$ & $1{,}627{,}126$ & $3.5083$ \\
210 & $0.6164 \pm 0.1829$ & $2{,}858{,}095$ & $2.8055$ \\
240 & $0.5970 \pm 0.1821$ & $2{,}508{,}457$ & $2.7380$ \\
270 & $0.6235 \pm 0.1840$ & $2{,}383{,}217$ & $2.8813$ \\
\textbf{300} & $0.5789 \pm 0.1839$ & $\mathbf{1{,}571{,}046}$ & $2.8778$ \\
330 & $0.6306 \pm 0.1859$ & $2{,}412{,}544$ & $2.3473$ \\
360 & $0.5939 \pm 0.1893$ & $2{,}159{,}588$ & $2.5695$ \\
390 & $0.6034 \pm 0.1831$ & $2{,}004{,}806$ & $2.4555$ \\
\textbf{420} & $\mathbf{0.5634 \pm 0.1916}$ & $1{,}801{,}407$ & $\mathbf{2.3217}$ \\
450 & $0.5945 \pm 0.1866$ & $2{,}292{,}435$ & $2.7691$ \\
480 & $0.6082 \pm 0.1934$ & $1{,}894{,}631$ & $2.6829$ \\
510 & $0.5851 \pm 0.2001$ & $2{,}371{,}493$ & $2.6039$ \\
540 & $0.5800 \pm 0.1878$ & $1{,}745{,}570$ & $2.3680$ \\
570 & $0.5977 \pm 0.1982$ & $2{,}042{,}874$ & $2.6032$ \\
600 & $0.5981 \pm 0.2035$ & $1{,}955{,}574$ & $2.4135$ \\
\bottomrule
\end{tabular}
\end{table}

Table~\ref{tab:full_cs2} reports LPIPS, FVD, and JEDi for \texttt{baseline\_v1} at each evaluated checkpoint on the CS2 test set (1,090 episodes). The epoch-510 checkpoint was unavailable due to a storage-retention error and is omitted rather than interpolated.

\begin{table}[h]
\centering
\caption{LPIPS, FVD, and JEDi for \texttt{baseline\_v1} at each evaluated checkpoint on the CS2 test set.}
\label{tab:full_cs2}
\begin{tabular}{cccc}
\toprule
Epoch & LPIPS $\downarrow$ & FVD $\downarrow$ & JEDi $\downarrow$ \\
\midrule
30  & $0.5734 \pm 0.1916$ & $4{,}313{,}603$ & $6.3680$ \\
60  & $0.5515 \pm 0.2077$ & $\mathbf{2{,}898{,}653}$ & $5.5596$ \\
90  & $0.5579 \pm 0.2081$ & $4{,}373{,}784$ & $4.0920$ \\
120 & $0.5203 \pm 0.2112$ & $3{,}411{,}944$ & $4.4658$ \\
150 & $0.5277 \pm 0.2075$ & $3{,}714{,}381$ & $3.5952$ \\
180 & $0.5040 \pm 0.2041$ & $3{,}113{,}235$ & $4.3480$ \\
210 & $0.5566 \pm 0.2196$ & $3{,}880{,}509$ & $3.6229$ \\
240 & $0.5004 \pm 0.2111$ & $3{,}821{,}083$ & $3.8476$ \\
270 & $0.4994 \pm 0.2075$ & $3{,}563{,}083$ & $3.3441$ \\
300 & $0.5185 \pm 0.2134$ & $3{,}369{,}055$ & $3.3322$ \\
330 & $0.5147 \pm 0.2204$ & $3{,}839{,}823$ & $2.8948$ \\
360 & $0.5073 \pm 0.2129$ & $4{,}124{,}846$ & $3.2223$ \\
390 & $0.4869 \pm 0.2075$ & $3{,}221{,}144$ & $3.0460$ \\
420 & $0.5169 \pm 0.2186$ & $4{,}084{,}299$ & $2.7355$ \\
450 & $0.5202 \pm 0.2233$ & $5{,}011{,}397$ & $2.6581$ \\
480 & $0.5081 \pm 0.2156$ & $4{,}119{,}733$ & $2.9661$ \\
540 & $0.4836 \pm 0.2102$ & $4{,}677{,}464$ & $2.5539$ \\
570 & $0.4814 \pm 0.2068$ & $3{,}995{,}518$ & $2.7391$ \\
\textbf{600} & $\mathbf{0.4785 \pm 0.2068}$ & $3{,}671{,}916$ & $\mathbf{2.3917}$ \\
\bottomrule
\end{tabular}
\end{table}

\section{Action-State Consistency Pipeline Example}
\label{app:strategy_spec}
The snippet below windows a source, measures pixel shift for the W control, scores each window, keeps only windows with at least one measured press, and samples a coverage-balanced curated set
 Importing \texttt{kuration.gameplay} registers \texttt{.pixel\_shift(...)}, which measures the optical-flow displacement behind every isolated key or mouse press (Section~\ref{sec:asc}), and the \texttt{action-state-consistency} scorer, which checks each window's own measured presses against a supplied baseline (an expected sign and magnitude range per control) and attaches a per-window anomaly count:

\begin{verbatim}
import kuration.gameplay  # registers pixel_shift and the scorer
from kuration import input

curated = (
    input({"clip1": rows}, schema=schema,
          fields={"action": action_columns}, time="t")
    .join_video({"clip1": "clip1.mp4"})
    .window(64, stride=32)
    .pixel_shift(controls=["W"])
    .score("aintropy/action-state-consistency",
           out="action_state_anomaly",
           baselines=baselines, controls=["W"])
    .filter("w_shift_measured", ">=", 1)
    .bin("w_shift_mean", bins=8)
    .sample("equal", n=100)
    .output("out/gameplay")
)
\end{verbatim}

Every segment in the output carries \texttt{action\_state\_anomaly}, the count of that window's own pixel-shift instances that violate the baseline, so a curation strategy can filter, bin, or weight on it directly.

\end{document}